\documentclass{article}

\usepackage{arxiv}
\usepackage[utf8]{inputenc} % allow utf-8 input
\usepackage[T1]{fontenc}    % use 8-bit T1 fonts
\usepackage{hyperref}       % hyperlinks
\usepackage{url}            % simple URL typesetting
\usepackage{booktabs}       % professional-quality tables
\usepackage{amsfonts}       % blackboard math symbols
\usepackage{nicefrac}       % compact symbols for 1/2, etc.
\usepackage{microtype}      % microtypography
\usepackage{lipsum}		% Can be removed after putting your text content
\usepackage{graphicx}
\usepackage[numbers]{natbib}
\usepackage{natbib}
\usepackage{doi}

\usepackage[font = scriptsize]{caption}
\usepackage{mwe}
\setcitestyle{square}
\usepackage{listings}
\usepackage{listings}
\usepackage[x11names]{xcolor}
\usepackage{subcaption}

\title{Parsing Musical Structure to Enable Meaningful Variations}

\author{%
  Maziar Kanani \\
  School of Computer Science\\
  University of Galway\\
  Galway, Ireland \\
  \texttt{m.kanani1@universityofgalway.ie}
  \\
  \And
  Seán O'Leary \\
  School of Computer Science \\
  TU Dublin\\
  Dublin, Ireland \\
  \texttt{sean.oleary@tudublin.ie} \\
  \AND
  James McDermott \\
  School of Computer Science\\
  University of Galway\\
  Galway, Ireland \\
  \texttt{james.mcdermott@universityofgalway.ie} \\
}

\begin{document}

\maketitle

\begin{abstract}
    This paper presents a novel rule-based approach for generating music by varying existing tunes.  We parse each tune to find the {\em Pathway Assembly} (PA)~\citep{marshall2019quantifying}, that is a structure representing all repetitions in the tune. The {\em Sequitur} algorithm~\cite{nevill1997identifying} is used for this. The result is a grammar. We then carry out mutation on the grammar, rather than on a tune directly. There are potentially 19 types of mutations such as adding, removing, swapping or reversing parts of the grammar that can be applied to the grammars. The system employs one of the mutations randomly in this step to automatically manipulate the grammar. Following the mutation, we need to expand the grammar which returns a new tune. The output after 1 or more mutations will be a new tune related to the original tune. Our study examines how tunes change gradually over the course of multiple mutations. Edit distances, structural complexity and length of the tunes are used to show how a tune is changed after multiple mutations. In addition, the size of effect of each mutation type is analyzed. As a final point, we review the musical aspect of the output tunes. It should be noted that the study only focused on generating new pitch sequences. The study is based on an Irish traditional tune dataset and a list of integers has been used to represent each tune's pitch values.
\end{abstract}

\section{Introduction}

Recently there has been a lot of research in the field of music generation as a research area in AI and computational creativity. There are many approaches to generate music, mainly rule-based and machine learning approaches. Allowing you to create complex structures by following a set of rules, rule-based systems have become a popular method for music generation. One area of rule-based music generation is the use of grammars. A grammar provides a systematic way of representing musical structures and generating music based on a set of rules. While some systems generate music from scratch, it is also interesting to generate music by variation from existing music, and that is our focus here.

An understanding of a piece's internal structure is necessary in order to manipulate it. By understanding its structure, we understand how a piece works and assume the composer intended it. One of the most critical principles is that the shortest representation of an object captures its internal structure best~\cite{kolmogorov-intro,kolmogorov1965three,grunwald-mdlintro,schmidhuber1997low}.

We use the concept of Pathway Assembly (PA) as the representation of musical internal structure~\cite{marshall2019quantifying}. In the theory of PA, the object is deconstructed into its irreducible parts, then evaluated in terms of how many steps are required to reconstruct it. We use the Sequitur algorithm as a method of capturing the PA structure~\cite{nevill1997identifying}.

\begin{figure}

  \centering
  \includegraphics[width=1\linewidth]{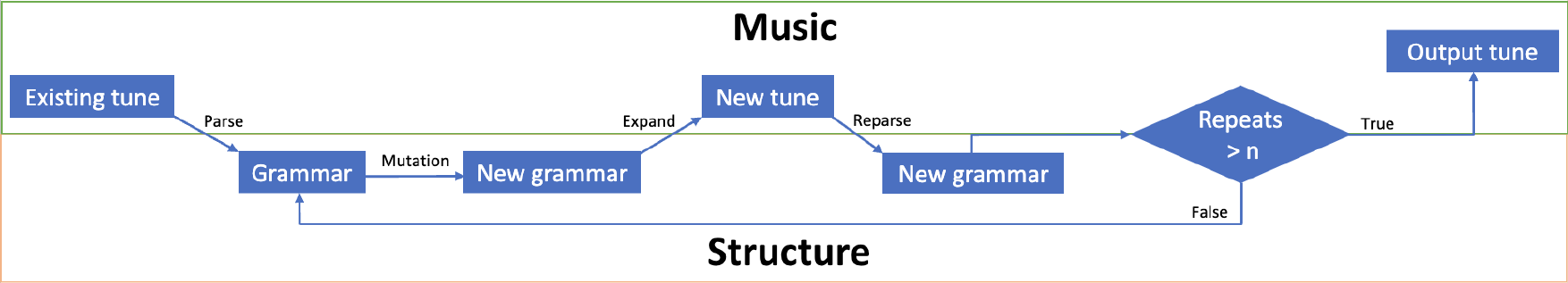}
  \caption{System diagram. Firstly, it receives an existing tune as a list of integers. By using Sequitur, it creates a grammar as the tune representation. Then a mutation is applied and the new grammar is created. The next step is to expand the new grammar. By expanding the grammar we have an updated list of integers that represent the new tune. After that, there is a reparse process and repeat decision discussed later in the paper. \label{fig:how}}
\end{figure}

As shown in Fig.~\ref{fig:how}, we next generate new variations of the music using 19 types of mutations to apply on the grammar, such as adding, removing, swapping or reversing some parts of the grammar. By mutating at the level of the structure, rather than the level of the tune, we achieve changes which are coherent across the tune, rather than breaking patterns. Mutating tunes directly (without considering their structure) is also possible, but the problem with that is that then the changes are completely random and there is a high probability of breaking the structure of the tune after a few mutations. In fact, the result is more like some random notes instead of a new tune. Mutating the grammar of tunes helps to keep, edit, or create new structures.

In this study we are focused on Irish folk tunes, with a tune represented as a list of monophonic pitch values. The method can be adapted for other styles of monophonic music as well. The study is restricted to generation of pitch values, not rhythm, harmony or other music features. The method can be extended to consider these features.

In the following, first we discuss background and related works to the study. Then we devote a section to introduce our methods and dataset. After that, we elaborate the mutations followed by their effect on a tune. Before concluding the paper, we discuss the results from a musical point of view.

\section{Background and Related Works}

In order to manipulate a piece of music, it is necessary to form an ``understanding" of its internal structure. It is an important principle that the shortest representation for an object is the one which best captures regularity in its structure. The Kolmogorov complexity of an object is defined as the length of the shortest program whose output is (a specification for) that object~\cite{kolmogorov1965three}. Kolmogorov complexity indicates how much information is contained in an object, such as a string of characters~\cite{kolmogorov-intro}. It can be said that it provides a formal and rigorous way of defining and quantifying the concept of randomness concept. However, Kolmogorov complexity is incomputable, and in Algorithm Information Theory (AIT)/Minimum Description Length (MDL) researchers are not concerned with actually finding the short programs that represent objects~\cite{grunwald-mdlintro}. MDL is a method based on Kolmogorov complexity for model selection~\cite{kolmogorov1965three, kolmogorov-intro, grunwald-mdlintro, chaitin1977algorithmic}. 

Furthermore, Schmidhuber~\cite{schmidhuber1997low} discusses the idea of ``low-complexity art" and how it relates to human perception. Schmidhuber argues that humans have a natural preference for patterns that can be described or compressed using simple rules, and that this preference extends to the art we create and appreciate. Essentially, we assume that the simplicity we perceive is intentional and reflects the designer's or artist's skill in creating a work that is both elegant and efficient. In~\cite{mccormack2021enigma} the authors examined the complexity of generative art and design by various measures. The authors argue that complexity is a fundamental aspect of both human creativity and the natural world, but it remains difficult to define and quantify. Most art is not too simple, and not too complex, but somewhere in between.

Forming a representation of the deliberately designed internal structure of a piece of music can be reduced (simplified) to finding structure in a sequence, a topic which has been studied e.g.~by Simon and Kotovsky~\cite{simon1963human} and by Hofstadter and colleagues~\cite{hofstadter1995fluid}. 

The concept of pattern recognition is fundamental in many areas of science. Ren et al.~\footnote{\url{https://www.music-ir.org/mirex/wiki/2019:Patterns_for_Prediction}} proposed a set of patterns for the 2019 Music Information Retrieval Evaluation eXchange (MIREX) pattern prediction competition. For this paper, a pattern is any sequence of notes which is repeated more than once. By considering patterns in structure of a tune, we can preserve, modify, or create new structures that enhance the overall composition~\cite{mcdermott2010higher}.

Loughran and O'Neill~\cite{loughran2020evolutionary} provide a comprehensive review of the application of Evolutionary Computation (EC) methods in algorithmic music composition. The review covers problem domains, methods, representations, and fitness measures utilized in studies spanning from the 1990s to 2020. The authors address the difficulty of measuring the "fitness" or aesthetic value of music, noting that while different fitness measures exist, none can independently, objectively, and reliably determine what constitutes good music, as musicality lacks a precise definition. The paper delves into the field of Computational Creativity (CC), examining the relationship between creativity and evolutionary methods, and addressing objections to computers being considered creative. 

One of the most significant previous works in the field of study is GenJam. In 1994, Biles~\cite{biles1994genjam} introduced a genetic algorithm-based model that is able to improvise jazz solos in real-time~\cite{biles1994genjam}. ``GenJam maintains hierarchically related populations of melodic ideas that are mapped to specific notes through scales suggested by the chord progression being played." A human mentor guides GenJam in improvising full-chorus solos and it "trades fours" in real time with humans in "chase" choruses. GenJam shows that GAs can be an effective tool for searching a constrained melodic space. Biles elaborates on the idea of ``musically meaningful operators", in particular musically meaningful mutations. His goal is to ensure that his mutation doesn't destroy musical structures. Our study attempts to address this concern by examining another approach.

Grammars have often been used in the context of representing art and music, especially for understanding and for manipulation, e.g.~grammatical (GE) evolution~\cite{o2003crossover,mcdermott2010higher}. In another work~\cite{loughran2018grammatical} Loughran discusses and contrasts the development of Composing Pony and GEChucK music generation systems based on GE. The main purpose of The Composing Pony is to investigate self-adaptive fitness in a move towards complete creative autonomy. The extended Composing Pony version produces both melody and accompaniment. This process generates an individual "<piece>," which comprises multiple musical events. Each of these events can consist of an individual note, a chord, a turn or an arpeggio. A chord is described by these attributes, along with the interval value between the root note and each upper note. Turns and arpeggios are described by specifying the number of steps, the size of each step, and the direction of the step progression. Interval sizes and step-sizes can be altered to favour harmonic or dissonant relationships, thus changing the likely harmonicity of the resulting melodies. This grammatical structure produces a population of individuals whose phenotypes can be played as MIDI messages~\cite{loughran2018grammatical}.

\cite{kaliakatsos2018generating} explores the possibility of conceptual blending by integrating higher-level features from drum rhythm data. Using this method, 32 features are extracted from the input rhythms. They used ``the blending of features" approach to ensure that they can make variations of a piece of music that still keep some musical structure, not just random changes.

There are some previous experiments in Irish Folk music generation such as ~\cite{sturm2022tradformer}. The authors describe a neural network architecture for modeling Irish and Swedish folk dance music, and for generating new examples.

\section{Methods and dataset}

\subsection{PA and Sequitur Algorithm }

The method that we chose to represent the structure of the tunes is {\em Pathway Assembly} (PA)~\cite{marshall2019quantifying}.  PA refers to the process of building complex structures or systems by assembling simple components or building blocks in a specific order or pattern. The idea behind PA is to identify the rules or constraints that govern the interactions between building blocks. By discovering the assembly process, it is possible to interpret or create a wide range of complex structures or systems with specific properties and functions. This approach has applications in many fields, including chemistry, biology, and engineering.

To find the PA for a string like 'abracadabra', firstly the list of basic components and the method of combination need to be determined. In this case \verb+{a, b, c, d, r}+ is the list of basic objects and objects can be combined by concatenation. Here is the PA:

%\begin{lstlistings}
\verb|{a, b, c, d, r} -> ca -> ab -> ra -> cad -> abra -> cadabra -> abracadabra|~\cite{marshall2019quantifying}
%\end{lstlistings}

"The PA index (PAI) of an object is the length of the shortest pathway to construct the object starting from its basic building blocks."~\cite{marshall2019quantifying}. PA for \verb+abracadabra+ is 7 as we can achieve the string in 7 steps.

We used the {\em Sequitur} algorithm~\cite{nevill1997identifying} as the algorithm to discover the tune structure (and calculate PAI). The Sequitur algorithm identifies hierarchical structure in sequences, such as natural language text or genetic code. Iteratively, the algorithm replaces repeated pairs of adjacent symbols with a new terminal or non-terminal symbol previously created and represents the pair. When all repeated pairs have been replaced, the sequence can be represented as a binary tree with the non-terminal symbols as internal nodes, and the original symbols as leaves. Sequitur has been shown to produce highly compressible representations of a variety of sequences, including DNA, natural language text, and computer programs~\cite{nevill1997identifying}. Here is the Sequitur output for the \verb+abracadabra+ example:

\begin{verbatim}
    p0 -> p1 c a d p1
    p1 -> p3 p2
    p2 -> r a
    p3 -> a b
\end{verbatim}

We can use the Sequitur output to calculate the PAI. PAI is the sum of the number of joins (or number of elements - 1) in the rules. In this example, \verb+p3+ has one join between the rules elements (\verb+a+ and \verb+b+). It is the same for \verb+p2+ and \verb+p1+ while it is 4 for \verb+p0+. The PAI is therefore 1 + 1 + 1 + 4 = 7, as above.

\begin{figure}
  \centering
  \includegraphics[width= 0.5\linewidth,trim={3cm 9cm 3cm 9cm},clip]{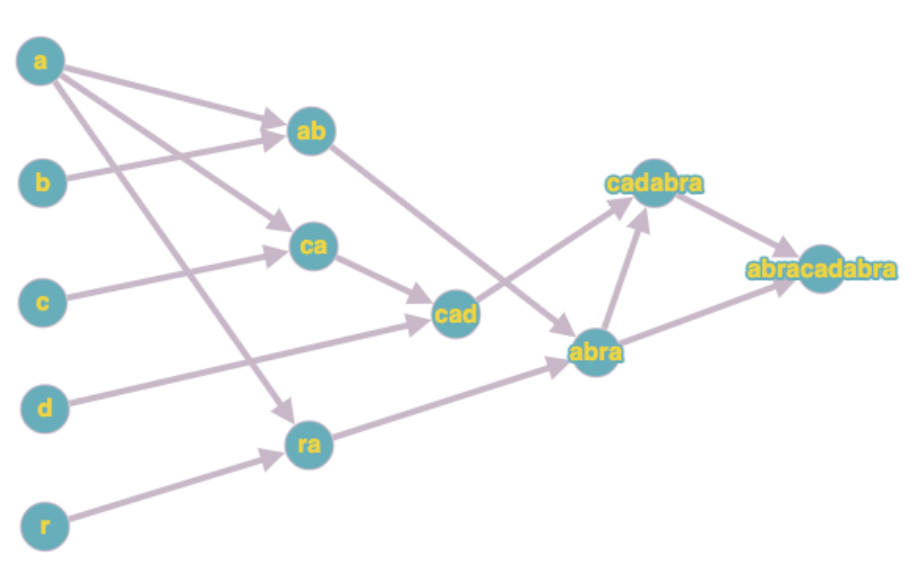}
  \caption{The PA for an object (abracadabra) is a DAG. The PAI is the number of compound objects in the DAG (here 7)~\cite{Kanani2023}. \label{fig:DAG}}
\end{figure}

To understand the relation between PA and Sequitur the grammar produced by the sequitur algorithm can be considered as a binary grammar: 

\begin{verbatim}
    P0 -> p0 p100 
    p1 -> p3 p2 
    P100 -> p101 p1 
    P101 -> p103 d 
    P103 -> ca
    p2 -> r a 
    p3 -> a b 
\end{verbatim}

This grammar corresponds directly to the DAG shown in Fig.~\ref{fig:DAG}. This DAG shows the same result in comparison to DAG of PA for the example.

In our study we represent a tune as a list of integers and use PA and Sequitur to find patterns as the representation of a tune structure.The system, uses Sequitur algorithm to parse tunes into grammar and re-parse them after mutations in the process of generating a new tune (see the example in Section~\ref{sec:example}). 

\subsection{Dataset} \label{sec:dataset}

The dataset of Irish traditional music used for our study is the {\em Ceol Rince na hÉireann} corpus, originally collected by Breathnach, preserved online by Bill Black, and processed by Diamond et al.~as part of the Polifonia project~\cite{diamond2022fonn}. The processed dataset comprises 1,195 Irish tunes, in MIDI, ABC, and raw pitch-integer notation. It is licensed under CC-by-v4\footnote{\url{https://github.com/polifonia-project/folk_ngram_analysis/blob/master/cre_corpus/README.md}}.

\subsection{Music Encoding} \label{sec: Music Encoding}
Before we start our analysis, we examined which representation of pitch is best for discovery of structure: absolute pitch values, or intervals. Surprisingly, pitch values produced the shortest grammar. Thus, we used pitch values to make a new tune.

\begin{figure}
    \centering
    \includegraphics[width = .7\linewidth]{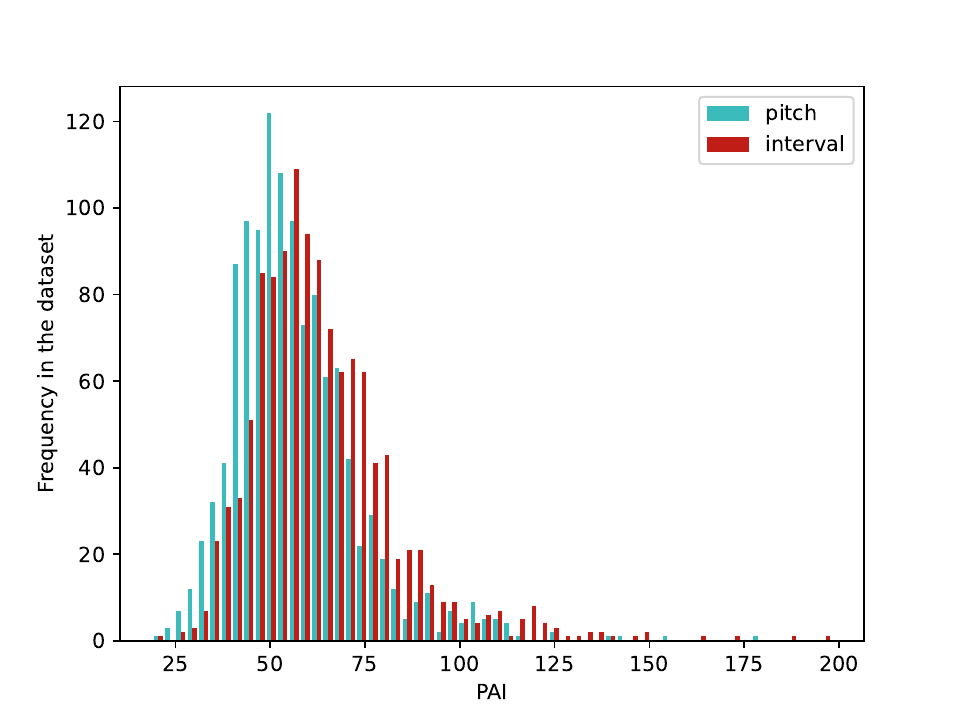}
    \caption{Shorter PAIs for pitch values compared to intervals}
    \label{fig:pitch v.s interval}
\end{figure}

\begin{figure}[htbp]
  \centering
  \makebox[\textwidth][c]{
  \begin{subfigure}[b]{0.45\textwidth}
    \includegraphics[width=\textwidth, trim={1cm 4cm 1cm 4cm}]{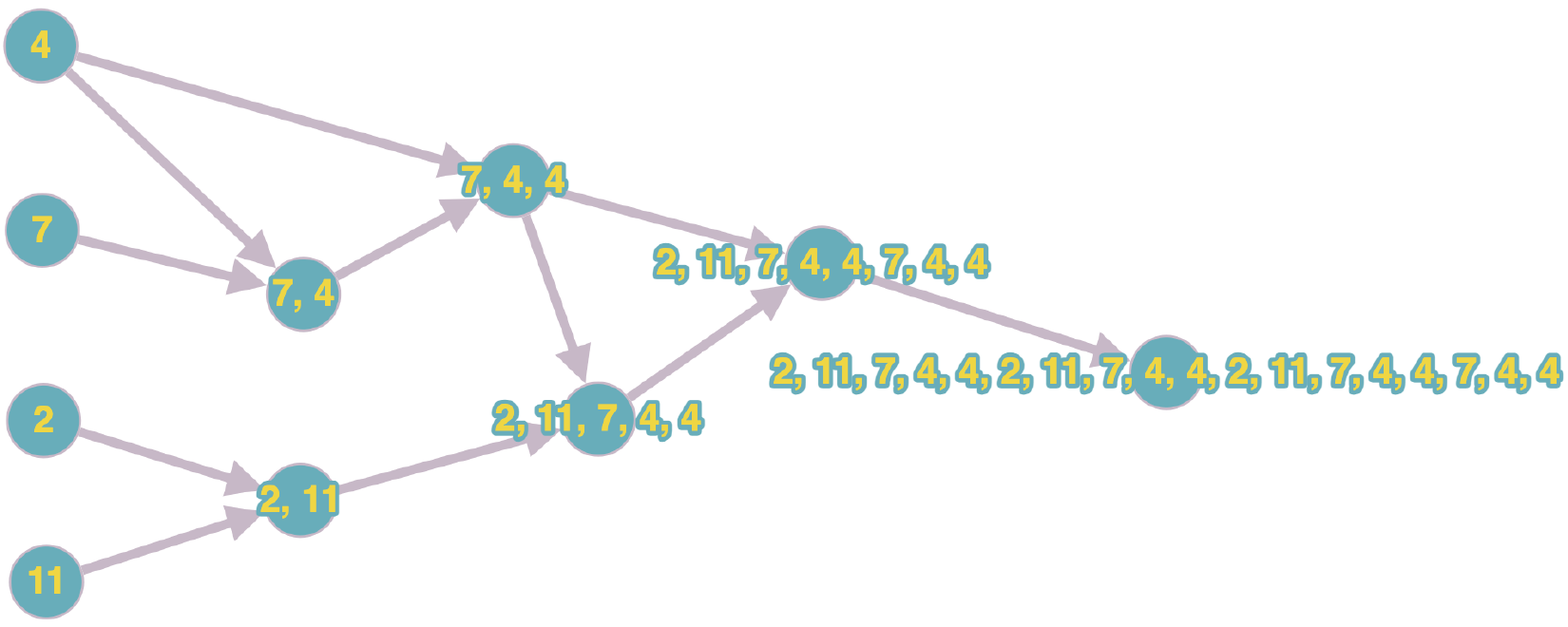}
    \caption{Pitch values PAI = 6 \\\ [2, 11, 7, 4, 4, 7, 4, 4, 2, 11, 7, 4, 4, 7, 4, 4]}
    \label{fig:graph2}
  \end{subfigure}
  \hfill
  \begin{subfigure}[b]{0.45\textwidth}
    \includegraphics[width=\textwidth, trim={2cm 4cm 1cm 4cm}]{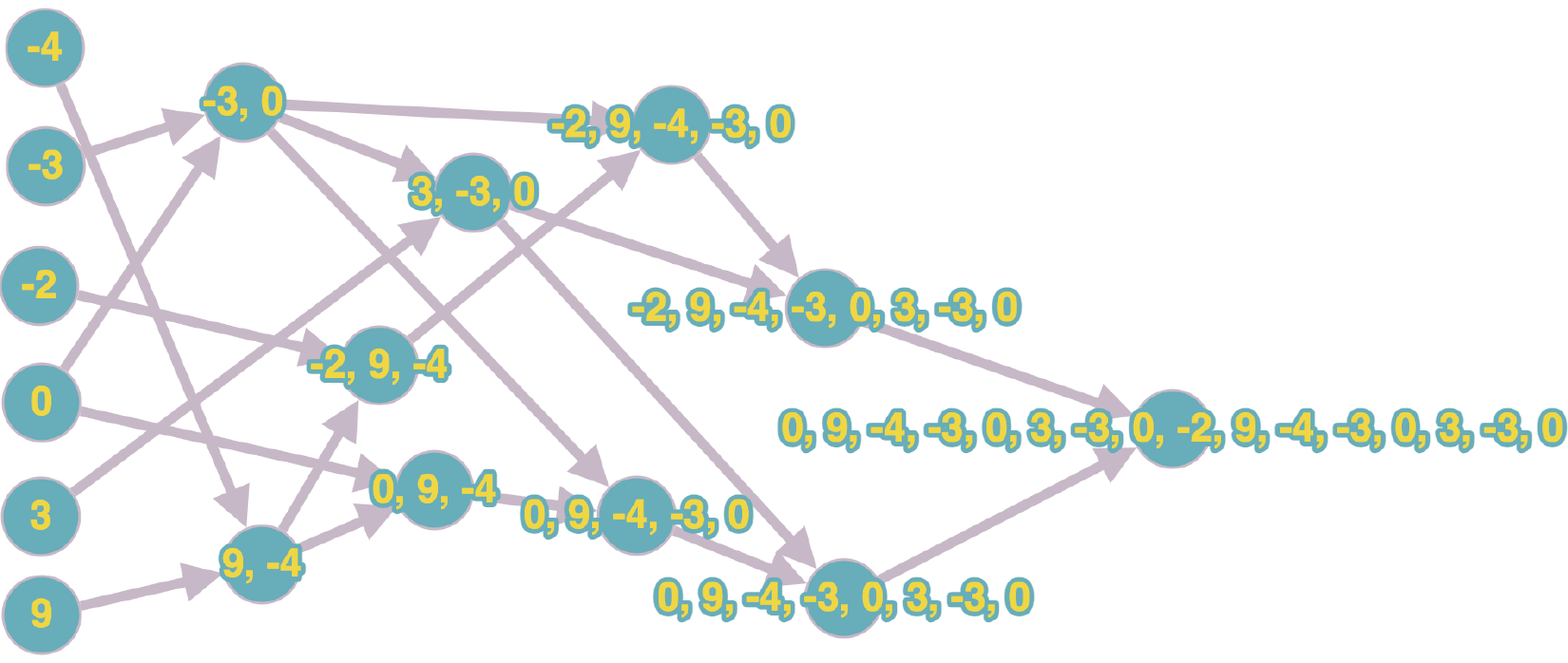}
    \caption{Intervals PAI = 10 \\\ [0, 9, -4, -3, 0, 3, -3, 0, -2, 9, -4, -3, 0, 3, -3, 0]}
    \label{fig:graph3}
  \end{subfigure}
  }
  \caption{Comparison of pitch values and intervals for the same part of a tune.}
  \label{fig:graphs}
\end{figure}

Fig.~\ref{fig:pitch v.s interval} illustrates the difference between the average of PAIs for tunes in the dataset, based on both pitch values and intervals. Clearly the average PAIs for pitch values are lower. Here is an example of PAIs computed for pitch values and intervals in a specific part of a tune that demonstrates the claim (Fig.~\ref{fig:graphs}):

\begin{verbatim}
    Pitch values: [2, 11, 7, 4, 4, 7, 4, 4, 2, 11, 7, 4, 4, 7, 4, 4]
    PAI = 6:
    
    {2, 4, 7, 11} -> 2, 11 -> 7, 4 -> 7, 4, 4 -> 2, 11, 7, 4, 4 ->
    2, 11, 7, 4, 4, 7, 4, 4 -> 2, 11, 7, 4, 4, 2, 11, 7, 4, 4, 2,
    11, 7, 4, 4, 7, 4, 4. 
    
    Intervals: [0, 9, -4, -3, 0, 3, -3, 0, -2, 9, -4, -3, 0, 3, -3, 0] 
    PAI = 10:
    
    {-4, -3, -2, 0, 3, 9} -> 9, -4 -> -3, 0 -> 0, 9, -4 -> 3, -3, 0 ->
    -2, 9, -4 -> 0, 9, -4, -3, 0 -> -2, 9, -4, -3, 0 -> 0, 9, -4, -3,
    0, 3, -3, 0 -> -2, 9,-4,-3, 0, 3, -3, 0 -> 0, 9, -4, -3, 0, 3, -3,
    0, -2, 9, -4, -3, 0, 3, -3, 0.  
\end{verbatim}

\section{List of mutations}

\begin{figure}
  \centering
  \includegraphics[width=1\linewidth]{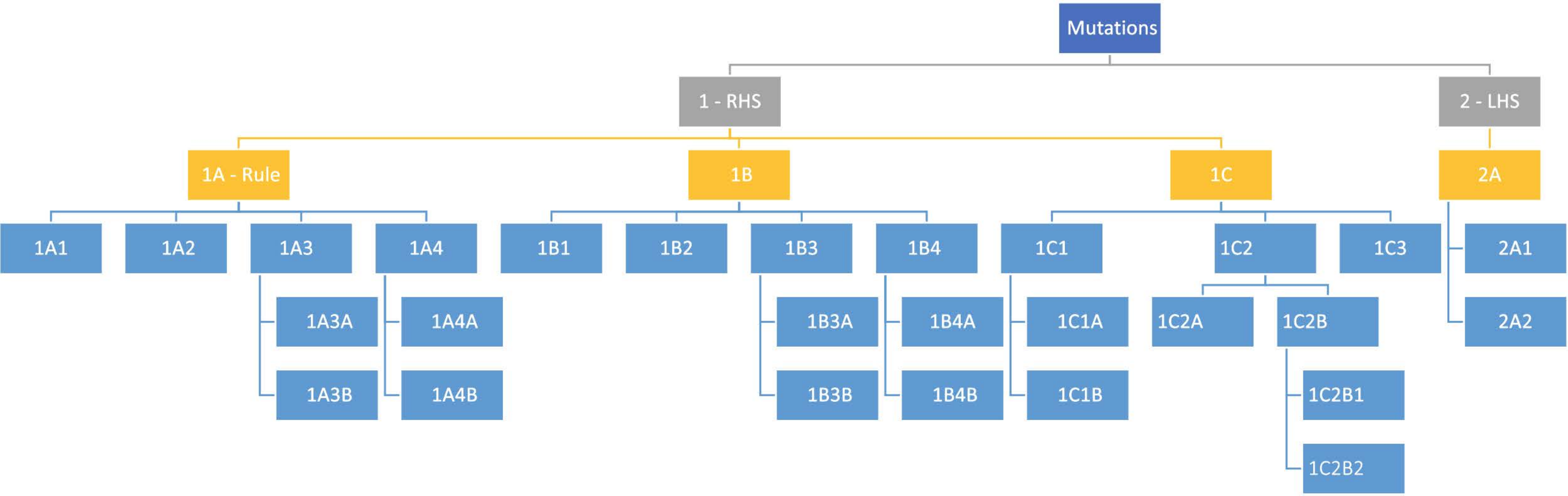}
  \caption{All the mutations and their connections\label{fig:mutations_list}}
\end{figure}

In this section, we discuss all types of possible mutations that can be applied to a grammar. Based on their location in the grammar, generally mutations can be divided into two main categories: the right-hand side (RHS) and the left-hand side (LHS). RHS mutation types can be divided into three sub-categories (Fig.~\ref{fig:mutations_list}).

RHS Mutations:

\begin{itemize}
\item Mutation 1 - 1A1:  Add an existing rule to another existing rule definition
\item Mutation 2 - 1A2: Remove a rule from another rule definition
\item Mutation 3 - 1A3A: Change the position of a rule - In a rule definition
\item Mutation 4 - 1A3B: Change the position of a rule – In two rule definitions
\item Mutation 5 - 1A4A: Swap the position of two rules – In a rule definition 
\item Mutation 6 - 1A4B: Swap the position of two rules – In two rule definitions 
\item Mutation 7 - 1B1: Add new number (choose a rule and an index in that rule's RHS, and then generate an integer from the list of integers and place that integer as a new note at that index)
\item Mutation 8 - 1B2: Remove a number (choose a rule and an integer on that, then removing the integer)
\item Mutation 9 - 1B3A: Change the position of a number – In a rule definition
\item Mutation 10 - 1B3B: Change the position of a number – In two rule definitions 
\item Mutation 11 - 1B4A: Swap the position of two numbers - In a rule definition 
\item Mutation 12 - 1B4B: Swap the position of two numbers – In two rule definitions 
\item Mutation 13 - 1C1A: Swap a rule and a number – In a rule definition 
\item Mutation 14 - 1C1B: Swap a rule and a number – In two rule definitions 
\item Mutation 15 - 1C2: Reverse the definition of an existing rule 
\item Mutation 16 - 1C3: Reverse the definition of an existing rule partially
\item Mutation 17 - 1D: Swap two rules' definitions (if other rules have different constraints, we might need more random choices to carry out one mutation)
\end{itemize}

LHS Mutations:
\begin{itemize}
\item Mutation 18 - 2A1: Define a new rule (based on the existing rules and numbers in the original tune)
\item Mutation 19 - 2A2: Remove an existing rule (we cannot remove \verb+p0+ - see the last paraghraph in Section \ref{sec:example})
\end{itemize}

\subsection{Example of how mutations work} \label{sec:example}
Here we explore how a single mutation can affect a tune. We begin by taking the first phrase of{\em A Trip to Galway}. Here we will work in pitch-class format for simplicity:

% \begin{verbatim}
%     [2, 11, 7, 4, 4, 7, 4, 4, 7, 11, 2, 1, 
%     2, 9, 6, 2, 2, 6, 2, 2, 6, 9, 2, 1,
%     2, 11, 7, 4, 4, 7, 4, 4, 7, 11, 2, 4,
%     6, 2, 11, 9, 6, 2, 4, 6, 4, 4]
% \end{verbatim}
\begin{lstlisting}[escapeinside={(*@}{@*)}]
    [2, 11, 7, (*@\colorbox{lightgray}{4, 4}@*), 7, (*@\colorbox{lightgray}{4, 4}@*), 7, 11, (*@\colorbox{lightgray}{2, 1,}@*)
    (*@\colorbox{lightgray}{2}@*), 9, 6, 2, 2, 6, 2, 2, 6, 9, (*@\colorbox{lightgray}{2, 1,}@*)
    (*@\colorbox{lightgray}{2}@*), 11, 7, (*@\colorbox{lightgray}{4, 4}@*), 7, (*@\colorbox{lightgray}{4, 4}@*), 7, 11, 2, 4,
    6, 2, 11, 9, 6, 2, 4, 6, (*@\colorbox{lightgray}{4, 4}@*)]
\end{lstlisting}

Then parsing it to obtain a short grammar, as below:

\begin{lstlisting}[escapeinside={(*@}{@*)}]
    p0 -> 2 p1 p2 9 p3 p3 6 9 p2 p1 2 4 p4 11 9 p4 4 6 p5
    p1 -> 11 p6 p6 7 11
    p2 -> (*@\colorbox{lightgray}{2,1,2}@*)
    p3 -> p4 2
    p4 -> 6 2
    p5 -> (*@\colorbox{lightgray}{4, 4}@*)
    p6 -> 7 p5
\end{lstlisting}

 To illustrate the effect of a mutation as an example, by using mutation 17 we swap the right-hand sides of \verb+p2+ and \verb+p5+. The changed grammar is:

%  \begin{verbatim}
 
%     p0 -> 2 p1 p5 9 p3 p3 6 9 p5 p1 2 4 p4 11 9 p4 4 6 p2
%     p1 -> 11 p6 p6 7 11
%     p2 -> 2 1 2
%     p3 -> p4 2
%     p4 -> 6 2
%     p5 -> 4 4
%     p6 -> 7 p2

% \end{verbatim}
\begin{lstlisting}[escapeinside={(*@}{@*)}]
    p0 -> 2 p1 p2 9 p3 p3 6 9 p2 p1 2 4 p4 11 9 p4 4 6 p5
    p1 -> 11 p6 p6 7 11
    p2 -> (*@\colorbox{lightgray}{4, 4}@*)
    p3 -> p4 2
    p4 -> 6 2
    p5 -> (*@\colorbox{lightgray}{2, 1, 2}@*)
    p6 -> 7 p5
\end{lstlisting}

After expanding the grammar we have the new list. The new list shows that the length of the original list has increased by 3 notes:

 % \begin{verbatim}
 
 %    [2, 11, 7, 2, 1, 2, 7, 2, 1, 2, 7, 11, 
 %    4, 4, 9, 6, 2, 2, 6, 2, 2, 6, 9, 4,
 %    4, 11, 7, 2, 1, 2, 7, 2, 1, 2, 7, 11, 
 %    2, 4, 6, 2, 11, 9, 6, 2, 4, 6, 2, 1, 2]

 % \end{verbatim}
\begin{lstlisting}[escapeinside={(*@}{@*)}]
    [2, 11, 7, (*@\colorbox{lightgray}{2, 1, 2}@*), 7, (*@\colorbox{lightgray}{2, 1, 2}@*), 7, 11, 
    (*@\colorbox{lightgray}{4, 4}@*), 9, 6, 2, 2, 6, 2, 2, 6, 9, (*@\colorbox{lightgray}{4,}@*)
    (*@\colorbox{lightgray}{4}@*), 11, 7, (*@\colorbox{lightgray}{2, 1, 2}@*), 7, (*@\colorbox{lightgray}{2, 1, 2}@*), 7, 11, 
    2, 4, 6, 2, 11, 9, 6, 2, 4, 6, (*@\colorbox{lightgray}{2, 1, 2}@*)]
\end{lstlisting}

\begin{figure}
  \centering
  \includegraphics[width=1\linewidth]{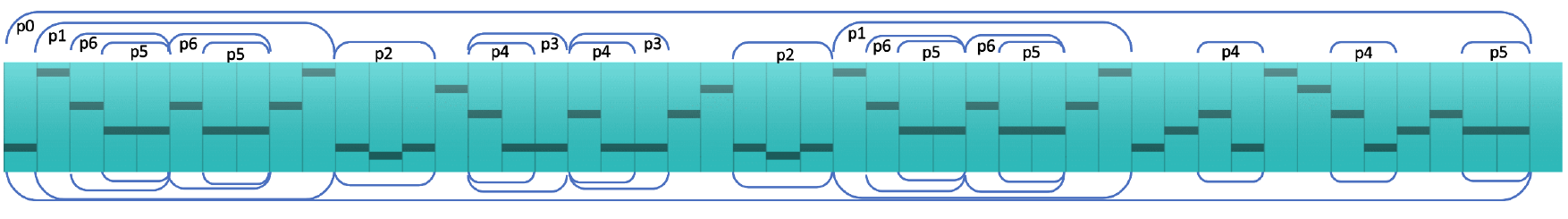}
  \includegraphics[width=1.062\linewidth]{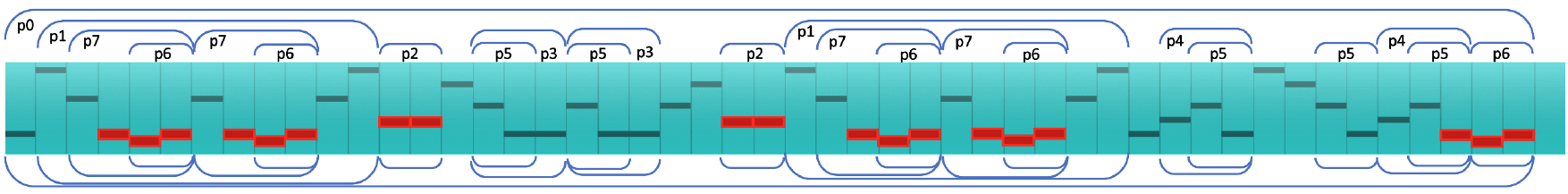}
  \caption{An example of how mutation works. After swapping the RHS of two rules, most of the structure is preserved. The new tune is re-parsed, which leads to (unimportant) re-numbering of the rules.\label{fig:mutation_example}}
\end{figure}

Fig.~\ref{fig:mutation_example} shows how the tune changed after the mutation. When the method adds or removes a note, it causes an offset for later notes. In most cases, tune and bar lengths change after applying our mutations. From another point of view, the notes after mutation become unaligned from the original alignment. In Fig.~\ref{fig:mutation_example} we can see the length has increased and is unaligned compared to the original version. There are three notes in each bar, beginning with the second note. For example, at the end of the first bar, we can see that this mutation has resulted in longer bars while bar 4 became shorter. The method keeps the length of each note for one beat and changes the length of bars and tunes.

After this process, the grammar may not obey the Sequitur principles (e.g., a rule may be defined even though it is not used). However, the grammar can still be expanded without problem. It means we need to firstly expand the grammar to produce the new tune, and then re-parse that to have a grammar which obeys Sequitur principles. That is the reason that in Fig.~\ref{fig:how} after achieving a new tune we have to reparse the new tune. After re-parsing the list, we obtain a new grammar:

 \begin{lstlisting}
 
    p0 -> 2 p1 p2 9 p3 p3 6 9 p2 p1 2 p4 11 9 p5 p4 p6 
    p1 -> 11 p7 p7 7 11 
    p2 -> 4 4 
    p3 -> p5 2 
    p4 -> 4 p5 
    p5 -> 6 2 
    p6 -> 1 2 
    p7 -> 7 2 p6
 \end{lstlisting}
 
This iterative process is repeated until the desired number of mutations is achieved. As a result, we get a new tune that is derived from the original. In fact, by applying the Sequitur algorithm, some parts of the structure of the tune are preserved. In the real system, we use pitch (not pitch class as in this example).

This illustration \ref{fig:mutation_example} demonstrates how a single mutation can have impact on the generated tune. There are multiple impacts in different places in the tune, and they are consistent with each other, rather than destroying patterns. While in this example the length of the musical phrase has increased, reducing its length after mutation is also possible. Additionally, the example shows how changes in the definitions of two specific rules have resulted in the tune. The changes are highlighted in red. The rule with three elements, \verb+p2+, replaces a rule with two elements, \verb+p5+, and as a result, there is an updated grammar with some parts in common with the original grammar.

In this study, to numerically compare the tunes pre- and post-mutation, the metric is "edit distance" (ED), also known as Levenshtein distance. We used ED because we want to calculate how much a tune has changed on the surface. This measure calculates the difference between two lists of strings by calculating the minimum number of single-character additions, deletions, and substitutions needed to transform one string into the other. For instance, when comparing "kitten" and "sitting", the ED is three, as three editing operations are required (i.e., 'k' to 's', 'e' to 'i', and adding 'g' at the end) to transform one word into the other.~\footnote{ Example from Wikipedia: \url{https://en.wikipedia.org/wiki/Edit_distance}} In the case of the example mutation in a phrase of {\em A Trip to Galway}, the ED between the initial list and the mutated list is 21. However, in general the amount of changes in the mutated tune could be significantly more or less than 21. Indeed, not only the type of mutation but also the grammar and the selected rule can have a considerable impact on the resulting ED. In this study \verb+p0+ is always the main rule and includes all the tune information in one rule. In other words, all other rules are connected to \verb+p0+, and by expanding \verb+p0+ we achieve the pitch value list of the tune. The main rule is followed by some other rules called primary rules. In general, primary rules are one to three rules following \verb+p0+ that provide a lot of information about the tune. Grammatical details are usually provided by the rest of the rules. Generally, primary rules are more substantial and have more connections with later rules, leading to more changes in the original tune when selected for mutation. The next section is devoted to exploring the effects of each type of mutation. Furthermore, there are more musicological metrics that we hope to adapt for future works.

\section{Mutation Effects} \label{sec:mutation_effects}

\begin{figure}
    \centering
    \includegraphics[trim={3cm 0 0 0}, clip, width=1.1\linewidth]{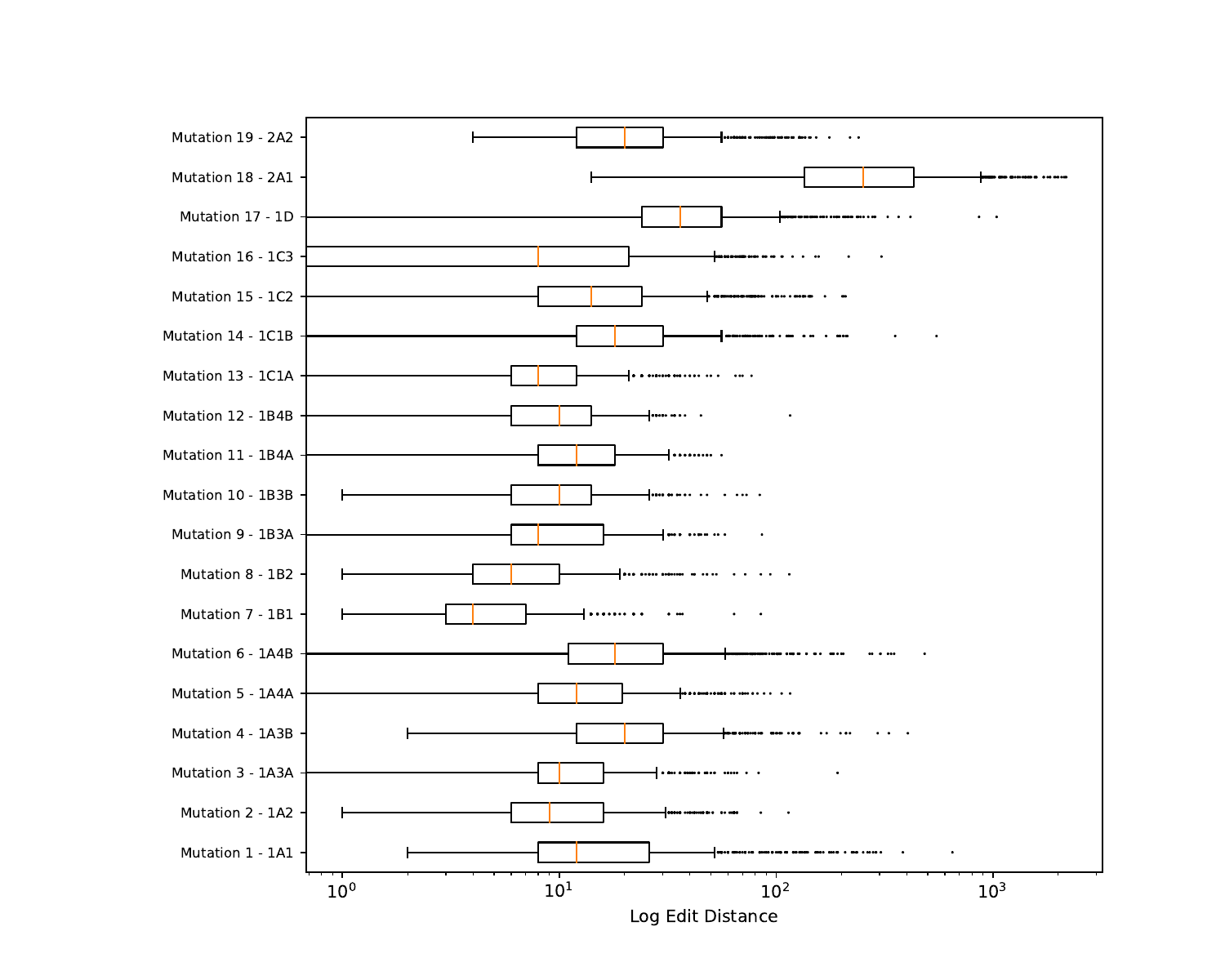}
    \caption{Box plot of running all the mutations on all the tunes once\label{fig:all_tunes}}
\end{figure}

Fig.~\ref{fig:all_tunes} displays the outcome of running each mutation once for each tune in the dataset independently. The running process randomly chooses a mutation type, and then the chosen mutation runs. We present one sample run on all the tunes. The presented log-scale ED is depicted for one of the runs for the dataset, revealing that mutation 18 has a significant impact and dominates over the other mutations. Following mutation 18, the deeper effects on tunes come from mutations in order 17, 19, 14, 6 and 4. It also shows ED averages are usually between 5 to 40 for most of the mutations. On the other hand, surface mutations are 7, 8, 9, 13 and 16. 

\begin{figure}
    \centering
    \includegraphics[trim={4cm 8cm 4cm 3cm}, clip, width=1\linewidth]{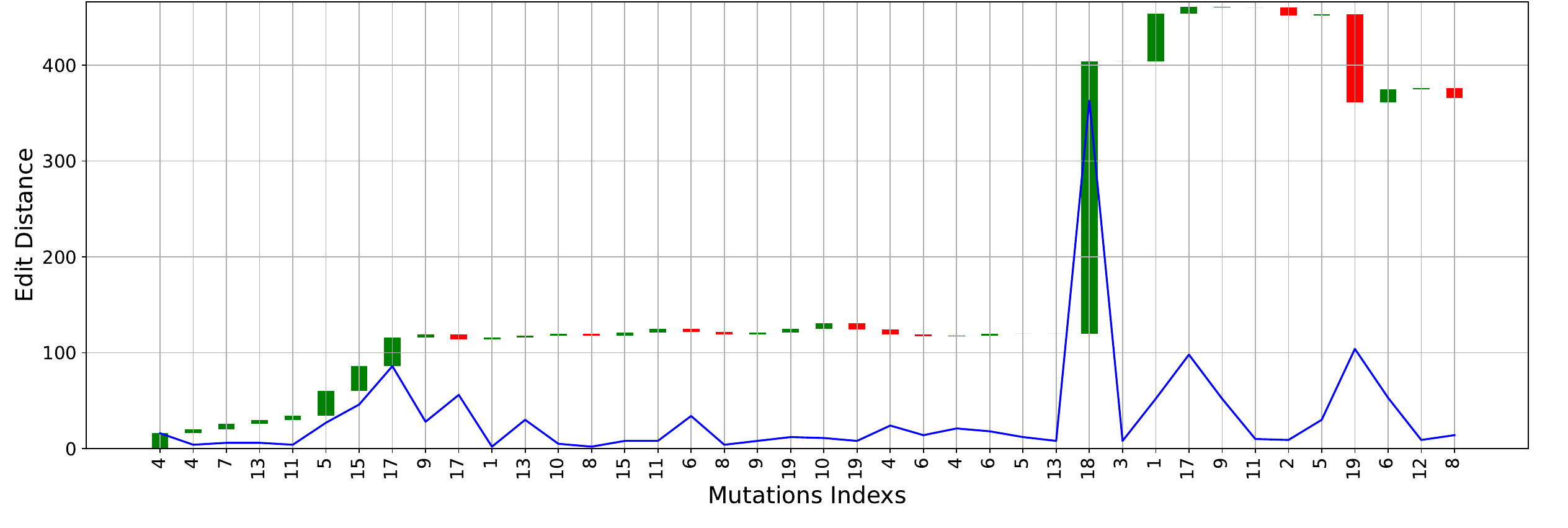}
    \caption{candlestick bar and line graph\label{fig:candlestick}}
\end{figure}

In Fig.~\ref{fig:candlestick} we can see an example of 40 successive random mutations chosen from the list of 19 mutations on the {\em A Trip to Galway} tune. We record how in each step one random mutation changes the original version. It is shown by green and red colors respectively for increasing or decreasing the ED versus the original (the candlestick graph). Furthermore, the line graph shows the ED between the successive versions of the tune. As the graph shows, mutation 18 has the largest effect. Sometimes mutation 19 has a relatively large effect. In this example, there is also significant  increases in ED caused by mutations 1, 4, 5, 15, 17. As compared to other mutations, they have significant impacts on the ED. The rest of the mutations show relatively slight changes in the tune. Furthermore, we can see that the ED tends to increase with the number of mutations.

\begin{figure}
    \centering
    \includegraphics[width=0.7\linewidth]{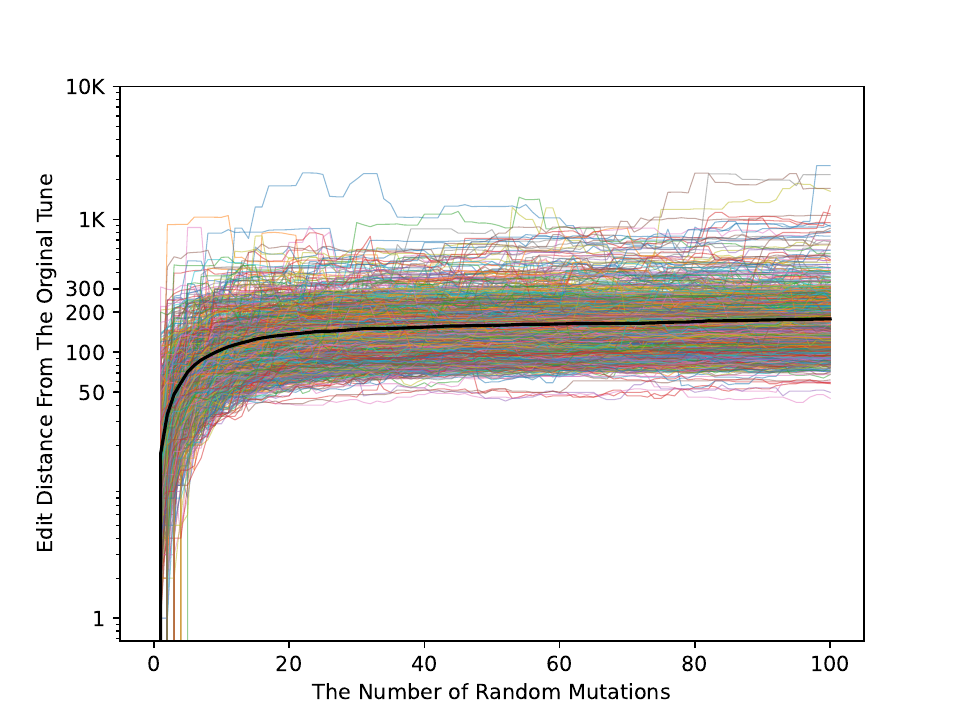}
    \caption{ED for running 100 random mutations for all the tunes and the ED average\label{fig:average_ED}}
\end{figure}

Considering mutation 18 dominates the results, it is excluded from the calculation of average EDs in the following analyses and also from generating new tunes. 
To further study the effect of multiple successive mutations, we are going to examine multiple mutations in succession. Fig.~\ref{fig:average_ED} shows the ED for 100 random mutations (from the list of 18 mutations) for all tunes in the dataset. The black line shows the average ED for all the tunes in each step. Seeing the graph, most tunes experience a slower increasing ED after 20 mutations, more like an exponential distribution pattern. After 20 mutations, the average ED slowly increases above 100, with an average of 178 for the dataset after 100 mutations. It is noteworthy that some tunes experience EDs exceeding 1000, mostly due to large increases in length. On the other hand, the final ED is less than 40 for some other tunes. Such outliers may occur due to the specific characteristics of individual tunes (e.g. mutating a very long rule in their grammar) or the effect of certain mutations on the grammar rules applied.

\section{Discussion}

\begin{figure}
  \centering
  \includegraphics[width=1\linewidth]{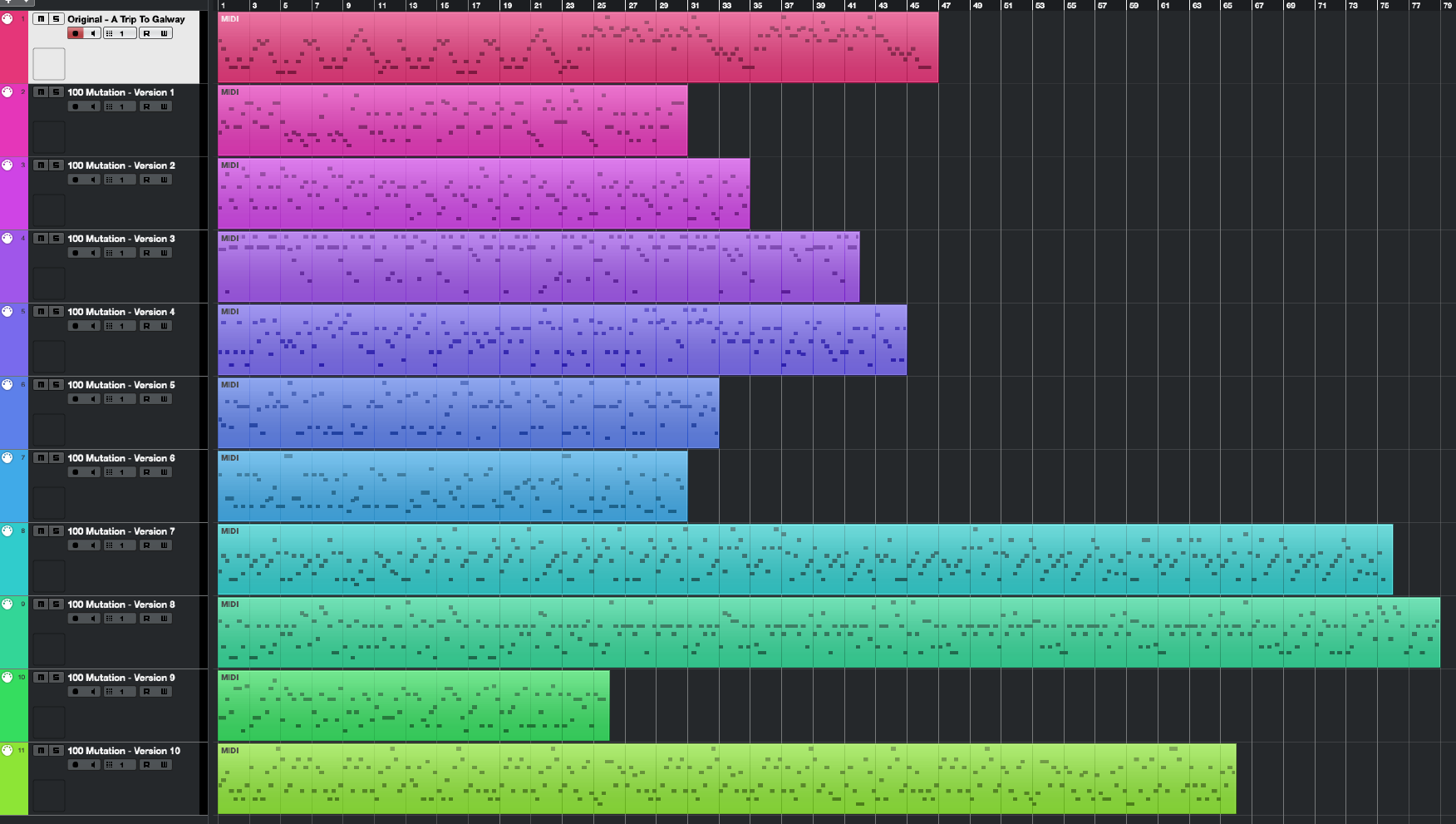}

  \caption{10 runs of the system on{\em A Trip to Galway}, 100 mutations per time\label{fig:runs}}
\end{figure}

We examined the outputs after a varied number of random mutations, ranging from a single mutation up to 100 mutations. The length of tunes is strongly connected with the types of mutations that randomly occur. The other critical factor is the rule in grammar that the randomly selected  mutation impacts. Therefore, we observed shorter, longer or nearly equal to the original tune lengths after 100 mutations (Fig.~\ref{fig:runs}).

We found some tunes hovering around a specific note in some cases. Also in some cases the system makes a pedal note during the tune. Both of these features give a more natural melodic sense in the novel tunes. 

It is important to note that mutations do not take into account the time signature and rhythm of a sequence. As a result, the created patterns can be longer or shorter than the beats in the original tunes. Taking rhythm into account in future work will allow more control over the high-level structure of the tune, making changes to musical patterns more evident to the listener.

In this study we focus on how to keep patterns and orders in pitch values. By "higher-level structure” we mean only the structure of tunes that includes those patterns and orders. The system can preserve that structure while changes at metric levels is probable. The mutated tunes use only notes that come from the original tune. The perceived tonic could change, but we have observed that this is not common.

The mutations do not consider cadences (perfect, plagal, imperfect or interrupted). Due to that, it is possible that the resulting tune will finish unexpectedly. In future works, it is worthwhile to add some constraints to avoid that. This will enable us to have an acceptable cadences both at the end of the phrases and the end of tunes.

\begin{figure}
  \centering
  \includegraphics[trim={4cm 4cm 4cm 3cm}, clip,width= 1\linewidth]{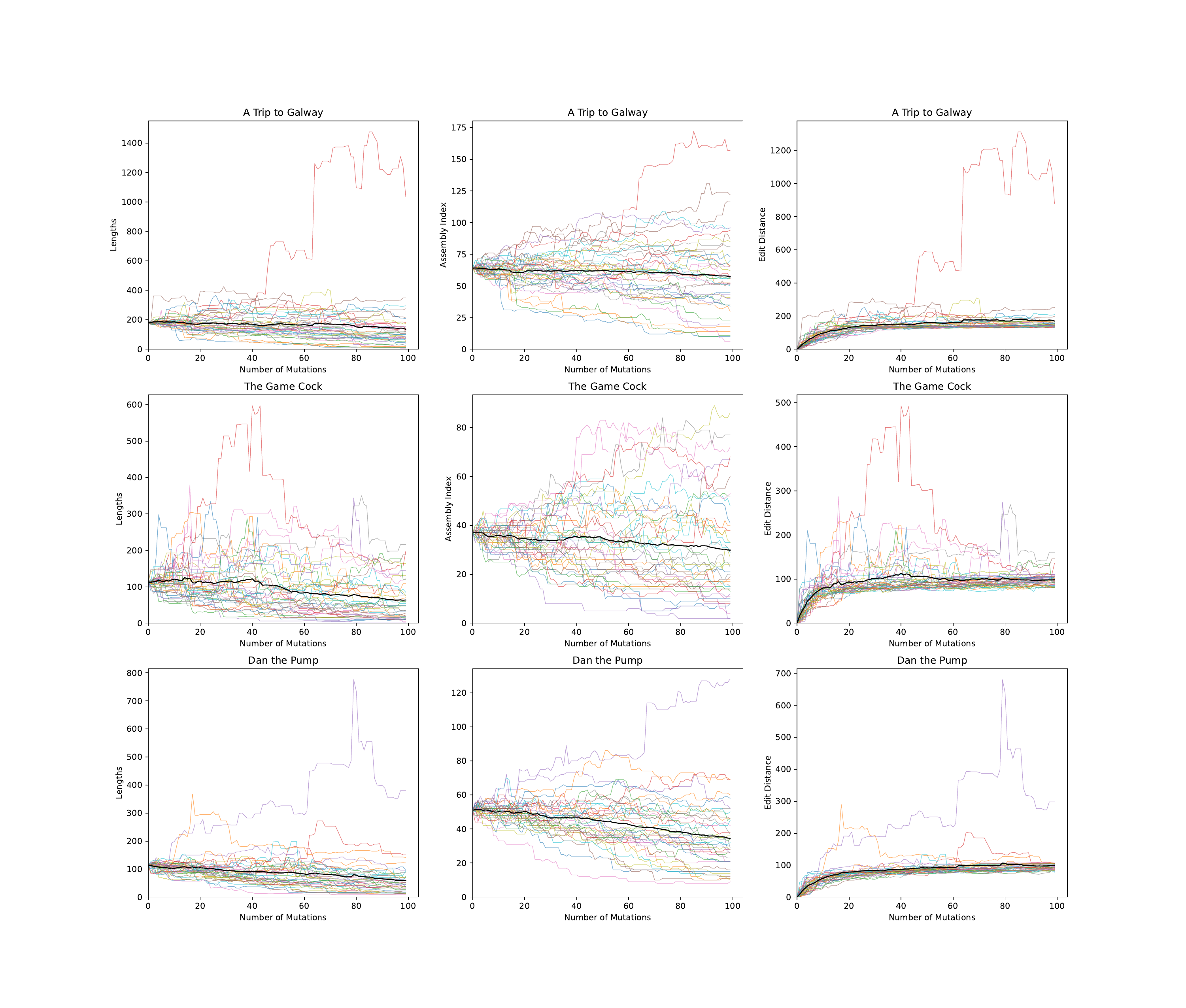}
  \caption{Length, ED, PA for three tunes over 100 mutations. The thick black line is the mean.\label{fig:multi_plot}}
\end{figure}

Fig.~\ref{fig:multi_plot} displays length, PA index (PA) and ED changes of three tunes over 100 mutations. The thick black lines indicate that the mean length and PA decrease after 100 mutations. After ignoring mutation 18 (2A1) we have three types of mutations that remove a part of the structure. In contrast, there are two adding types of mutations. Having one more removing mutation type than adding mutation type is why we see a decrease on the average length. 

Although the complexity decreases slightly, it is balanced by the decrease in length. Thus we might say that the "complexity per unit length" remains approximately the same. This tends to show that the mutations defined over grammars have the desired effect, that is preserving the amount of structure. They do not gradually increase entropy in the tunes, as a mutation defined over pitch values directly might~\cite{mcdermott2010higher}.

Previously, we talked about the ED average of all the tunes. The plot shows that the behaviour is the same for these three tunes as the behaviour for the entire dataset~\footnote{More examples in more details: \url{https://ufile.io/bdl7dk3j}}.

\begin{figure}
  \centering
  \includegraphics[width=1\linewidth]{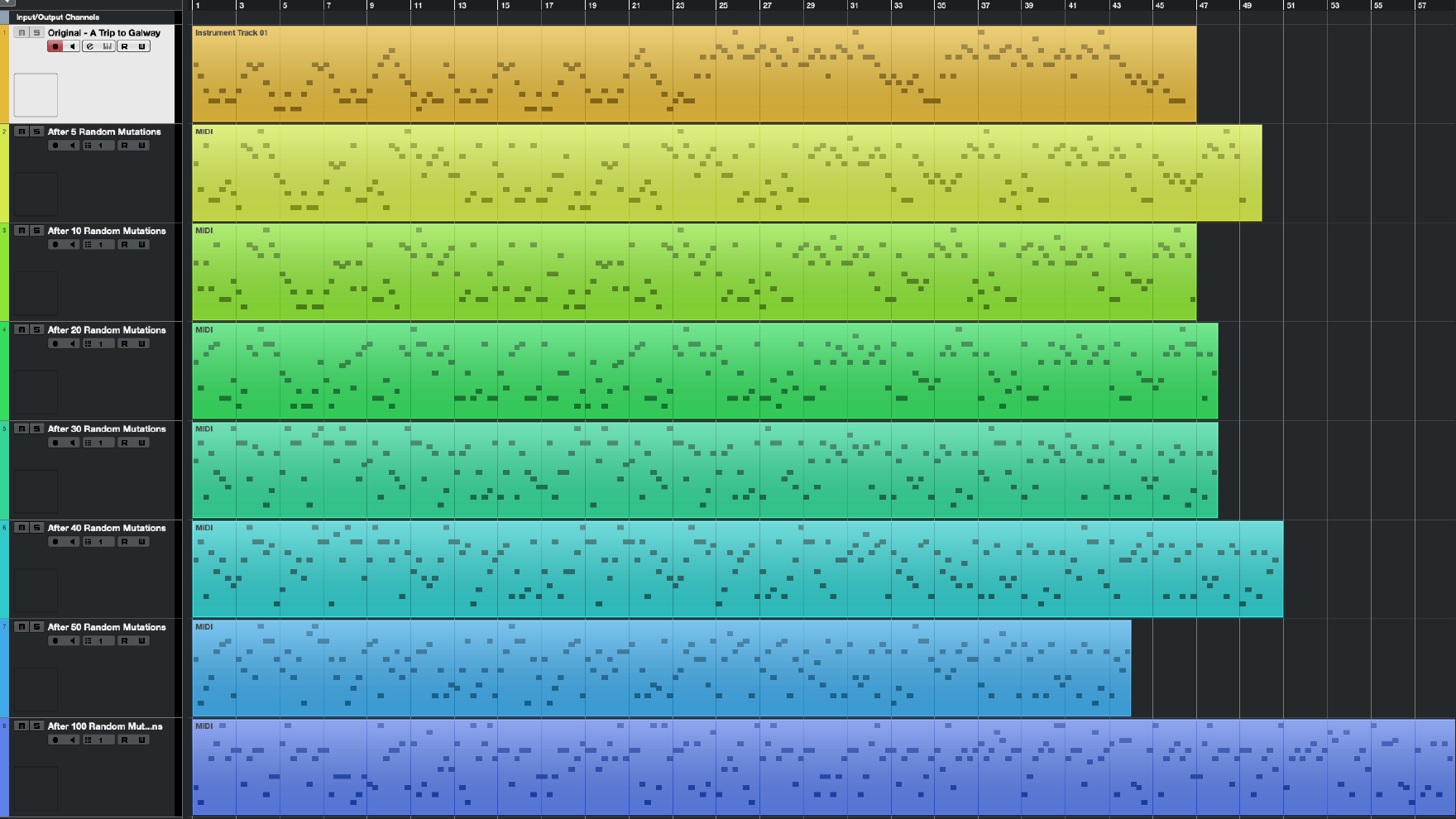}

  \caption{{\em A Trip to Galway} after 5, 10, 20, 30, 40, 50 and 100 mutations\label{fig:step_by_step}}
\end{figure}

Figure~\ref{fig:step_by_step} shows piano rolls of the original tune ({\em A Trip to Galway}) and mutated tunes after 5, 10, 20, 30, 40, 50 and 100 mutations. It illustrates the evolution of the tune with the number of mutations. Usually after 5 mutations we still  a strong similarity to the original tune, but with some noticeable changes. From approximately 10 to 30 mutations we hear a new tune that is still recognisably connected to the original version. After 40 to 50 mutations the tune is less connected to the original and after 100 mutations it can sometimes be said that only the modal key is in common with the original version.

\section{Conclusion and future works}

Based on the idea of finding the structure of music and parsing the musical structure, the Pathway Assembly and Sequitur algorithm are adapted to obtain for each tune in our corpus a deterministic grammar (the PA object) which is a representation of that tune’s musical structure. These grammars include rules created by patterns and numbers. 19 types of mutation on grammars are introduced in this paper. The system uses mutations to make changes to the rules by manipulating numbers and grammars. Mutations are characterized by adding, removing, swapping, or reversing numbers and patterns in grammar rules. After that we demonstrate how these mutations affect tunes. 

We have found both in numerical experiments and through listening to the results that, as expected, this method helps to keep, modify, or create new patterns which tend to preserve the overall level of complexity in the tune, rather than introducing randomness without pattern.

In future works, we will try to constrain the algorithm to give more musical control over mutations, addressing problems like missing cadences or patterns which are out of sync. The Pathway Assembly representation is sufficient to represent many types of musical patterns, but not all. In future work, we will investigate more powerful representations, such as programs in a Turing-complete language, to capture a wider range of patterns.

\section{Ethics statement}
In the Irish oral tradition, some tunes have authors and some do not, but the community regards authenticity and provenance as important. Many in the community see themselves not only as individual musicians but as keepers of the tradition. Many in the community have reacted negatively to modern AI music tools which synthesize new tunes based on a corpus drawn from the tradition~\footnote{See e.g.~\url{https://thesession.org/discussions/47876}. {\em The Session} is probably the central online repository of tunes in the Irish traditional style, including newly-composed ones, and has now banned AI-generated tunes.}. Our system works differently: each new tune is a variant of a single, specific existing tune. This difference is important because the existing tune and its author (if any) are not hidden and their contribution is not minimised. Of course, these considerations do not prevent a malicious actor from generating new tunes and claiming them as hand-written.

Moreover, {\em our goal is not to create new tunes which ``fool'' the listener}. In fact, the tunes created by our system go outside the normal parameters of the tradition in clear-cut ways, as discussed. This gives an interesting, novel artistic effect rather than mere imitation of the corpus.

\section{Acknowledgments}
This work was conducted with the financial support of the Science Foundation Ireland Centre for Research Training in Digitally-Enhanced Reality (d-real) under Grant No. 18/CRT/6224.

\bibliography{refs}
\bibliographystyle{ieeetr}
\end{document}